%% file: main.tex
\documentclass[10pt,twocolumn,letterpaper]{article}

\usepackage{cvpr}              
\usepackage{xcolor,colortbl}


\definecolor{cvprblue}{rgb}{0.21,0.49,0.74}
\definecolor{pred}{RGB}{200, 1, 80}
\definecolor{grey}{rgb}{0.9, 0.9, 0.9}

\usepackage{multirow}
\usepackage{colortbl}
\newcommand{\ours}{LAIN\xspace}
\usepackage[accsupp]{axessibility}  

\usepackage[pagebackref,breaklinks,colorlinks,allcolors=cvprblue]{hyperref}

\def\paperID{*} 
\def\confName{CVPR}
\def\confYear{2025}

\title{ Locality-Aware Zero-Shot Human-Object Interaction Detection}

\author{Sanghyun Kim \hspace{0.8cm}  Deunsol Jung \hspace{0.8cm}  Minsu Cho \vspace{1.5mm}\\
Pohang University of Science and Technology (POSTECH), South Korea \\
{\tt\small \{sanghuyn.kim, deunsol.jung, mscho\}@postech.ac.kr
}
\\\small
\href{http://cvlab.postech.ac.kr/research/LAIN}{\url{http://cvlab.postech.ac.kr/research/LAIN}}
}

\begin{document}
\maketitle

\input{sections/0_abstract}    
\input{sections/1_introduction}
\input{sections/2_related_work}

\input{sections/3_method}
\input{sections/4_experiments}
\input{sections/5_conclusion}

{
    \small
    \bibliographystyle{ieeenat_fullname}
    \bibliography{main}
}


\end{document}

%% file: sections/0_abstract.tex
\begin{abstract}
Recent methods for zero-shot Human-Object Interaction~(HOI) detection typically leverage the generalization ability of large Vision-Language Model (VLM), \textit{i.e.,} CLIP, on unseen categories, showing impressive results on various zero-shot settings.
However, existing methods struggle to adapt CLIP representations for human-object pairs, as CLIP tends to overlook fine-grained information necessary for distinguishing interactions.
To address this issue, we devise, \ours, a novel zero-shot HOI detection framework designed to enhance the locality and interaction awareness of CLIP representations.
The locality awareness, which involves capturing fine-grained details and the spatial structure of individual objects, is achieved by aggregating the information and spatial priors of adjacent neighborhood patches.
The interaction awareness, which involves identifying whether and how a human is interacting with an object, is achieved by capturing the interaction pattern between the human and the object.
By infusing locality and interaction awareness into CLIP representations, \ours captures detailed information about the human-object pairs.
Our extensive experiments on existing benchmarks show that \ours outperforms previous methods in various zero-shot settings, demonstrating the importance of locality and interaction awareness for effective zero-shot HOI detection.

\end{abstract}

%% file: sections/1_introduction.tex
\section{Introduction}

The task of Human-Object Interaction (HOI) detection aims to localize human-object pairs and recognize the interactions between them in a given image, \textit{i.e.,} identifying a set of HOI instances (human, object, interaction). 
HOI detection is useful for a wide range of computer vision applications, including image retrieval~\cite{Wu_2022_CVPR,yoon2021image,gordo2017beyond} and image captioning~\cite{Wu_2022_caption,yao2018exploring,herdade2019image}, where a comprehensive understanding of human-object relationships is essential.
Although significant advances have been made recently, conventional HOI methods~\cite{li2024logic,lei2023ada,Kim_2023_CVPR} have primarily been relying on fully supervised learning, limited to identifying predefined HOI categories. 
Given that humans interact with objects in a compositional way, it is costly and impractical to collect annotations for all possible HOI categories, limiting their ability to identify novel HOI categories not present in the training set.
\begin{figure}
     \centering
         \centering
         \scalebox{0.98}{\includegraphics[width=\linewidth]{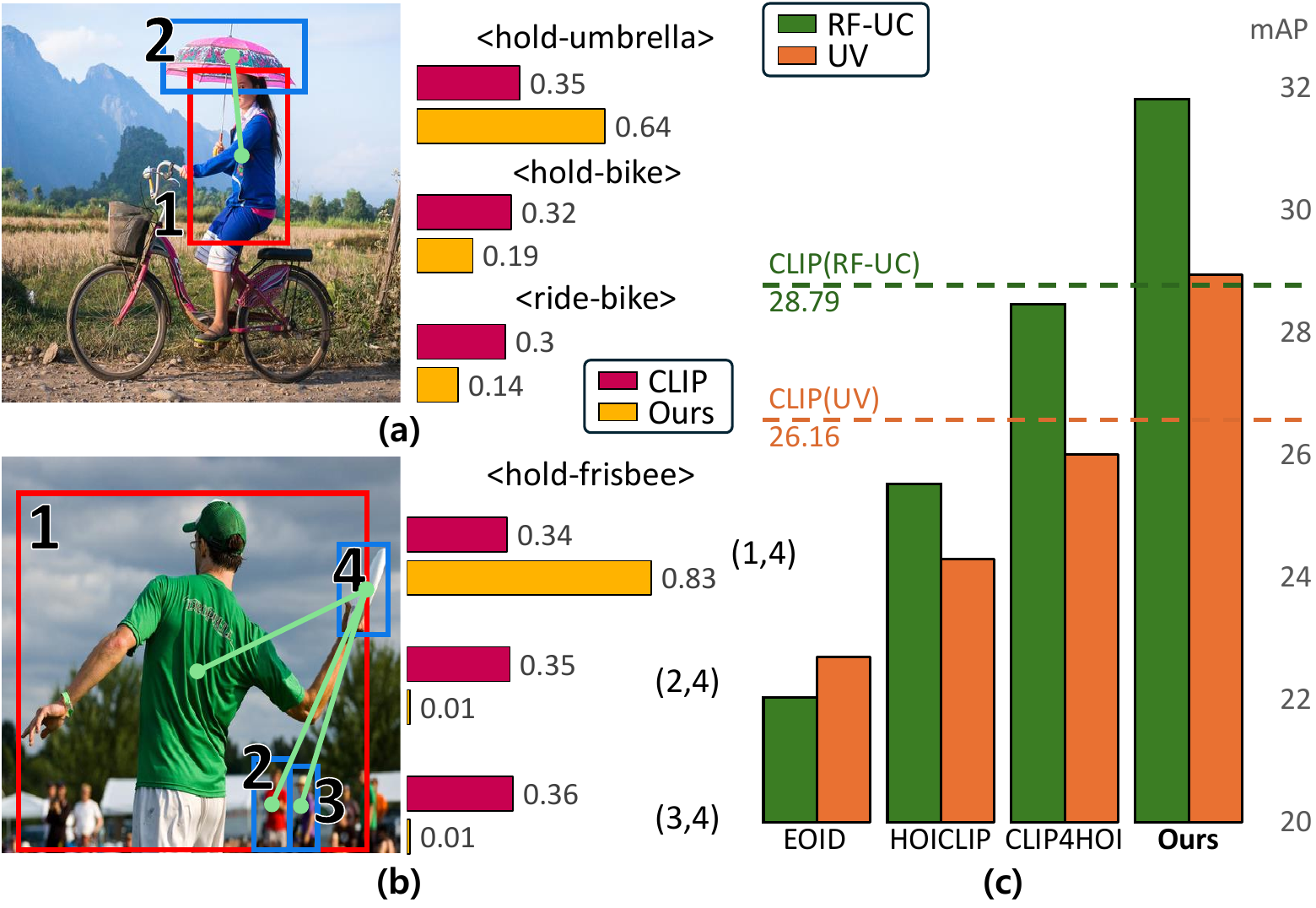}}
    \vspace{-2mm}
    \caption{(a)-(b): Since CLIP primarily encodes global information, it struggles to capture the fine-grained details required to accurately identify interactions within human-object pairs. (c): When existing methods adapt CLIP representations to zero-shot HOI detection, this limitation hinders CLIP's generalization, and results in degraded performance which is even lower than CLIP's original zero-shot performance in UC-RF and UV settings.
    }
     \vspace{-6mm}
    \label{fig:teaser}
\end{figure}
Recently, strong generalization ability of CLIP on unseen categories, which stems from contrastive image-level pre-training on large-scale data, has inspired the development of a zero-shot HOI model that leverages this capability to recognize unseen HOI categories.

While existing methods~\cite{ning2023hoiclip,wu2023eoid,mao2024clip4hoi} for zero-shot HOI detection have achieved strong performance by leveraging CLIP representations, the domain gap between the image-level pre-training task and the region-level task poses the challenges in adapting CLIP representations to zero-shot HOI detection.
Since CLIP predominantly encodes global information~\cite{zhong2022regionclip,zhou2023zegclip}, it often fails to extract fine-grained information about individual objects.  This hinders the HOI model from capturing whether and how the person interacts with the object.
As shown in Figure~\ref{fig:teaser}~(a),
CLIP assigns a high confidence score to the interaction `ride-bike' even though the human-object pair \textit{(1,2)} does not focus on the bike region, indicating that CLIP has limited capacity to capture fine-grained object details.
As a result, the human-object pair \textit{(1,4)} has a similar confidence score for the unseen HOI category `hold-frisbee' compared to others, \textit{i.e.,} \textit{(2,3)} and \textit{(3,4)}, struggling to distinguish the interactive pair in Figure~\ref{fig:teaser}~(b).
These issues weaken the generalization ability of CLIP when existing methods attempt to adapt its representations for zero-shot HOI detection, resulting in lower zero-shot performance compared to CLIP's original results as shown in Figure~\ref{fig:teaser}~(c).

To address the challenges mentioned above, we introduce a novel zero-shot HOI detection framework, dubbed \textbf{L}ocality-\textbf{A}ware \textbf{I}nteraction \textbf{N}etwork~(\ours), that learns locality-aware interaction via Locality Adapter~(LA) and Interaction Adapter~(IA).
The LA extracts locality-aware features from image patch tokens by considering visual context of neighboring regions and spatial priors, and then infuses them back into the image patch tokens.
The IA leverages the locality-aware patch tokens to update human-object tokens by performing interaction reasoning between human and object regions, resulting in interaction-aware human-object tokens.
In this manner, the LA provides fine-grained details and spatial structure for individual objects, enabling the IA to perform effective contextual reasoning. The IA complements CLIP representations by incorporating fine-grained details, providing a relational context that cannot be captured by locality awareness alone.
By incorporating locality and interaction awareness, which play complementary roles, each layer of~\ours effectively captures detailed information for the human-object pair, facilitating the adaptation of CLIP representations to zero-shot HOI detection.

To demonstrate the effectiveness of our proposed method, we conducted extensive evaluations on two public benchmarks, HICO-DET~\cite{hico} and V-COCO~\cite{vcoco}. The experimental results show that \ours outperforms the previous methods for zero-shot detection across all zero-shot settings, demonstrating the robust generalization ability of our approach in zero-shot scenarios.
Our ablation studies demonstrate the importance of locality and interaction awareness for zero-shot HOI detection.

Our contribution can be summarized as follows: 
\begin{itemize}
\item We propose the Locality-Aware Interaction Network (\ours), which incorporates a Locality Adapter~(LA) and an Interaction Adapter (IA).
\item By enriching CLIP representations with locality and interaction awareness, \ours~effectively captures the fine-grained details about human-object pairs.
\item Extensive experiments demonstrate that LAIN achieves outstanding zero-shot performance, achieving a new
state-of-the-art.
\end{itemize}

%% file: sections/2_related_work.tex
\section{Related work}
\subsection{Human-Object Interaction~(HOI) Detection}
Conventional HOI detection methods can be roughly divided into two categories: two-stage and one-stage methods.
Two-stage methods~\cite{zhang2022upt,zhang2022exploring,gao2020drg,gao2018ican,li2020IDN,qi2018dpnn,ulutan2020vsgnet,wang2020hetero,zhang2021spatially,hou2020vcl} first detect humans and objects using a pre-trained detector~\cite{ren2015faster,detr}.
After constructing all possible human-object pairs based on the detection results, these pairs are fed into an interaction classifier.
To generate discriminative features for classifying the interaction of a human-object pair, they incorporate additional information~\cite{li2019tin,liu2020consnet,gupta2019no} and perform relational reasoning on a graph structure~\cite{wang2020hetero,ulutan2020vsgnet,zhang2021spatially}.
In contrast to two-stage methods that follow a sequential cascade to determine the interaction between the human-object pairs, one-stage methods~\cite{fang2021dirv,liao2020ppdm,kim2020uniondet} concurrently detect individual instances, pair the human-object instances, and classify interactions.
Inspired by the transformer-based detector, \ie, DETR~\cite{detr}, where each query learns to detect an object, recent one-stage methods~\cite{tamura2021qpic,zhou2022distr,Kim_2023_CVPR,li2024logic} have adopted transformer-based structures, where each query predicts a (human, object, interaction) triplet.
Despite their promising results, these approaches heavily rely on full annotations with predefined HOI categories, which makes them impractical for handling unseen HOI categories.

\subsection{Vision tasks with CLIP}
CLIP~\cite{CLIP} is a multimodal framework that adopts contrastive learning to jointly train image and text encoders on large-scale image-text pairs found from the web.
By leveraging vision and language knowledge pre-trained on large-scale data, CLIP has significantly improved the zero-shot capabilities of models across various downstream tasks, including out-of-distribution  detection~\cite{wang2023clipn,esmaeilpour2022zero} and segmentation~\cite{zhou2023zegclip,dong2023maskclip}.
However, CLIP struggles to align local image regions with text descriptions since CLIP was trained by aligning whole images with their corresponding text descriptions in a common embedding space, thus producing suboptimal results on region-level tasks~\cite{zhong2022regionclip,dong2023maskclip,wu2023clipself,zhou2023zegclip,ma2022visual}. 
\begin{figure*}[t!]
   \centering
   \includegraphics[width=0.93\linewidth]{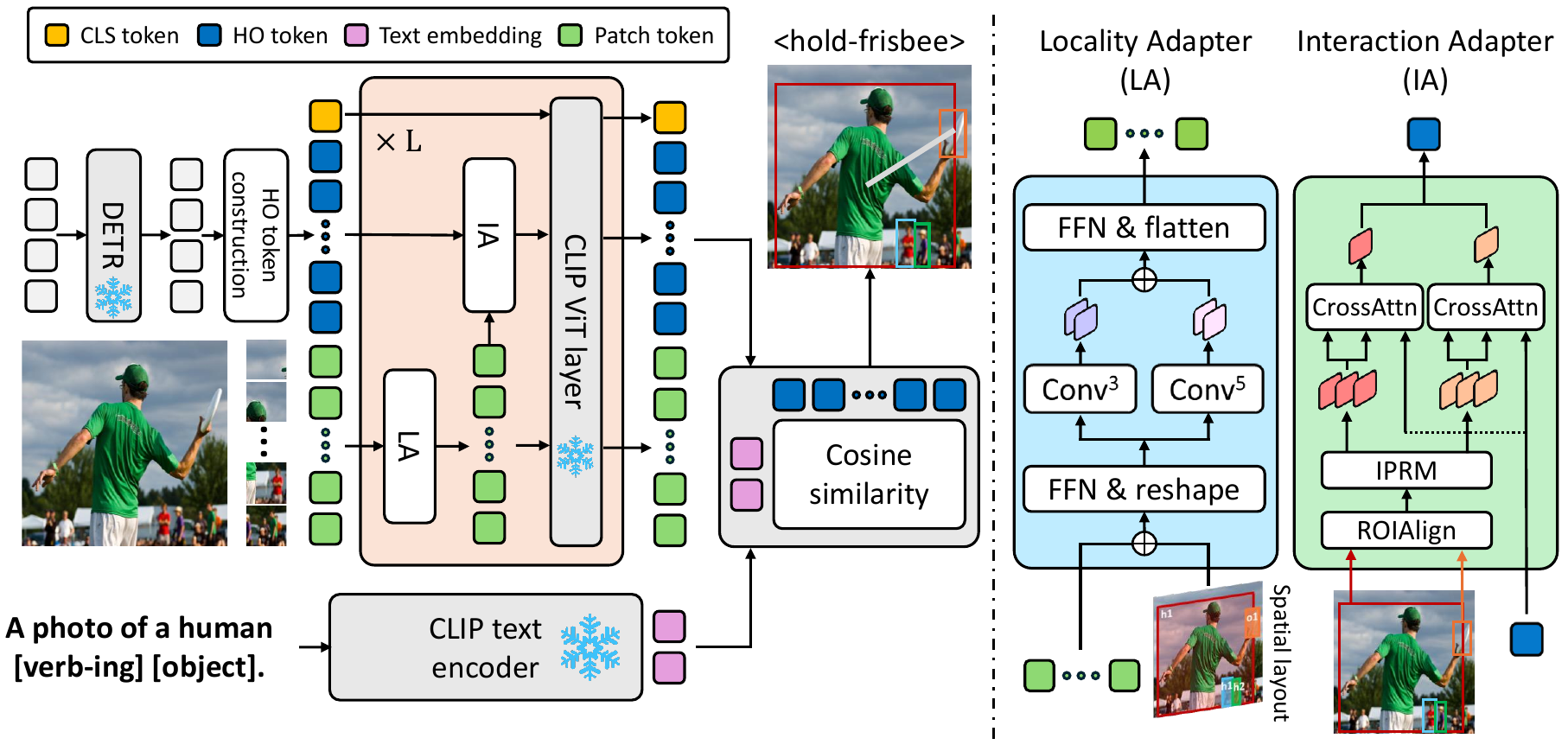}
   \vspace{-4mm}
   \caption{The overall architecture of \ours. All valid human-object pairs are constructed and embedded into HO tokens based on detection results from a pre-trained DETR~\cite{detr}. Image patch tokens are passed through the Locality Adapter (LA), which infuses locality awareness into each patch token. The updated patch tokens and HO tokens are then passed through the Interaction Adapter (IA), which enhances each HO token with interaction awareness. The HO, $\textrm{CLS}$, patch tokens are subsequently refined by the frozen $l$-th ViT layer of the CLIP~\cite{CLIP} visual encoder. After repeating this process for $L$ layers, HOI scores are computed by measuring the cosine similarity between the HO tokens and text embeddings extracted from CLIP text encoder.}
   \vspace{-5mm}
\label{fig:overall_archi}
\end{figure*}
To mitigate the issue, recent work~\cite{zhong2022regionclip,dong2023maskclip,wu2023clipself,chen2022vision,liu2021swin} learns a locality in the CLIP or ViT structure by training on large scale of the region-text pairs~\cite{zhong2022regionclip,chen2022vision,dong2023maskclip}, and introduces attention mechanism to capture the information of local regions~\cite{hassani2023neighborhood,liu2021swin}.
However, they require additional pre-training stages to adapt the downstream task, and require modifications to the network structure which limits the utilization of CLIP’s pre-trained knowledge.

\subsection{Zero-shot HOI detection}
Zero-shot HOI detection aims to detect both seen HOI categories available during training and unseen HOI categories that do not appear during training.
Previous work~\cite{hou2021affordance,liu2020consnet,hou2021detecting,hou2020vcl} mainly adopts compositional learning, which disentangles HOI representation into object and interaction features.
Although the disentangled features enable the model to recognize unseen combinations, these methods are limited to combinations where either the object or the verb is not shown in training, since it is infeasible to learn the disentangled features of unseen objects or verbs.
In light of the success of CLIP~\cite{CLIP} on various zero-shot settings, recent work~\cite{liao2022gen,wu2023eoid,wan2024exploiting} focuses on transferring CLIP knowledge via the teacher-student architecture. 
However, since they transfer knowledge only about seen classes, they tend to be biased toward seen category samples.
Furthermore, since this knowledge, \ie, CLIP scores, only conveys which interaction occurs, the model struggles to learn fine-grained local details such as human attire.
To enhance transferability, prior work~\cite{mao2024clip4hoi,ning2023hoiclip,lei2025exploring,lei2023ada} directly leverages CLIP representations. 
ADA-CM~\cite{lei2023ada} uses object queries from DETR as global object priors to inform all objects in the image, primarily providing global context to CLIP rather than local details of individual objects.
CMMP~\cite{lei2025exploring} uses spatial priors to model plausible spatial configurations, focusing only on spatial relationships rather than the relational context of an HO pair.
BCOM~\cite{wang2024bilateral} incorporates additional knowledge extracted from the detector backbone along with CLIP representations to better capture the small-scale interaction.

In contrast, our method enhances CLIP representations by integrating locality awareness for fine-grained object details and interaction awareness for the relational context of the HO pair, enabling better adaptation of CLIP representations to zero-shot HOI detection.

%% file: sections/3_method.tex
\section{Method}
\subsection{Overview}
Given an input image $I$, the goal of zero-shot HOI detection is to predict all HOI instances, including those belonging to HOI categories that are unseen during training. 
Formally, $i$-th HOI instance is defined \textit{4-tuple} $(b^h_i,b^o_i,c_i,a_i)$, where $b^h,b^o\in \mathbb{R}^4$ represent the bounding box coordinates of the human and object, respectively.
$c_i \in \mathbb{O}$ and $a_i \in \mathbb{V}$ represent the object class and the interaction class that occurs between a human and an object, \ie verb, where $\mathbb{O}=\{o_1, \cdots , o_{N_o}\}$ is a set of objects and  $\mathbb{V}=\{v_1, \cdots , v_{N_v}\}$ is a set of interactions. 
$N_o$ and $N_v$ are the number of object and interaction classes, respectively.
Under the zero-shot settings, the model is trained only on the samples from seen HOI categories $\mathbb{C}_{\textrm{seen}}=\mathbb{C} \backslash \mathbb{C}_{\textrm{unseen}}$, where $\mathbb{C}=\{(o_i,v_j)|o_i\in\mathbb{O}, v_j \in \mathbb{V}\}$ is the set of all possible HOI categories.

The overall architecture of \ours~is illustrated in Figure~\ref{fig:overall_archi}.
To detect all possible HOI instances, we first detect the objects in the input image $I$ using a pre-trained detector, \ie, DETR~\cite{detr}.
Based on the detection results, we exhaustively construct all valid human-object pairs and generate HO tokens which are used to classify the interaction for each corresponding human-object pair.
The HO tokens, along with the \textrm{[CLS]} token and the image patch tokens, are then passed through the CLIP visual encoder to aggregate visual information for each human-object pair.
We attach the Locality Adapter (LA) and Interaction Adapter (IA) to the front of each CLIP visual encoder layer, to inject locality and interaction awareness into the CLIP representations, respectively.
After repeating this process for $L$ layers, the HOI score for each HO token is computed by measuring the similarity to text embeddings of HOI categories.

\subsection{HO Token Construction}
\label{sec:HO_token_construction}
In this section, we introduce HO tokens for HOI detection, which aggregate and convey contextual information from an input image similar to the $\mathrm{[CLS]}$ token in CLIP.
Each HO token is constructed based on detection results to determine the interaction of the corresponding human-object pair.

Specifically, following the two-stage methods~\cite{zhang2022upt, mao2024clip4hoi}, given an image $I$, we use an off-the-shelf object detector, \ie, DETR~\cite{detr}, to obtain detection results $\{(b_i, c_i, s_i, g_i)\}^{N_{\mathrm{det}}}_{i=1}$, where $b_i\in\mathbb{R}^4$, $c_i \in \mathbb{O}$, $s_i \in \mathbb{R}^1$ and $g_i \in \mathbb{R}^{D_\mathrm{det}}$ represent the bounding box, object class, confidence score, and object feature, respectively.
$N_\mathrm{det}$ and $D_\mathrm{det}$ denote the number of detected objects and dimension of the object feature, respectively.
Based on the detected results, HO tokens $T \in \mathbb{R}^{N_\mathrm{pair} \times D_\mathrm{clip}}$ for all valid human-object pairs are constructed as follows:
\begin{align}
 \mathrm{idx} &= \{(u,v)~|~u \neq v,~c_u = \text{`}\text{human}\text{'} \}, \\ 
 T_i &= \mathrm{FFN}([g_u;g_v]), \text{where}~(u,v) = \mathrm{idx}_i, 
 \label{eq:HOtoken}
\end{align}
where $N_\mathrm{pair}$ is the number of all valid human-object pairs and $D_\mathrm{clip}$ is the feature dimension of the CLIP. 
The HO tokens $T$ are concatenated with the $\textrm{[CLS]}$ token and image patch tokens $F$, and then passed through the CLIP visual encoder composed of $L$ layers to aggregate the visual information:
\begin{equation}
    [T_{(l)}, \mathrm{cls}_{(l)}, F_{(l)}] = \mathcal{V}_{(l)}([T_{(l-1)};\mathrm{cls}_{(l-1)};F_{(l-1)}),
    \label{eq:vit}
\end{equation}
where $\mathrm{cls} \in \mathbb{R}^{1 \times D_\mathrm{clip}}$ and $F \in \mathbb{R}^{HW\times D_\mathrm{clip}}$ denote $\textrm{[CLS]}$ token and image patch tokens, respectively, and $\mathcal{V}_{(l)}$ indicates the $l$-th layer of the CLIP visual encoder. 
$H$ and $W$ are the height and width of the feature map before flattening.
For simplicity, we omit the layer index $l$ in the following sections. 

\subsection{Locality Adapter}
To determine the interaction between an HO pair, it is crucial to recognize fine-grained details of the individual object.
For example, if the model learns during training that a human is wearing a helmet in the seen HOI category `ride-bike', this fine-grained information can help to identify the unseen HOI category `ride-snowboard'.
However, while CLIP representation effectively captures global information, it lacks the ability to capture fine-grained local details in specific regions of the image~\cite{dong2023maskclip,wu2023clipself,zhou2023zegclip}, \ie, locality awareness.
To mitigate this, a Locality Adapter (LA) enhances locality awareness of CLIP by updating each patch token with aggregated information from neighboring tokens.
Specifically, we first reshape the flattened patch tokens $F$ to their original shape, and then project them through a Feed-Forward Network (FFN) to obtain the $\tilde{F} \!\in\! \mathbb{R}^{H \times W \times D_a}$, where $D_a <\!\!< D_{\mathrm{clip}}$.
Then, we construct the spatial layout embedding \( L_{i,j} = \text{FFN}([b_t; c_t; e_t]) \) according to the detection results, where \( t \) denotes the index of detected object corresponding to the position \((i,j)\), and \([\cdot;\cdot]\) indicates the concatenation operation.
Here, \( b_t \), \( c_t \), and \( e_t \) represent the box coordinates, confidence score, and object text embedding extracted from the CLIP text encoder, respectively.
Subsequently, the layout embedding $L\in \mathbb{R}^{H \times W \times D_a}$ is embedded into $\tilde{F}$
to provide spatial prior of entire objects in $I$ as follows:
\begin{align}
    \hat{F} = \textrm{LN}(\textrm{FFN}(\tilde{F} + L )),
    \label{eq:LA_feat}
\end{align}
where $\textrm{LN}$ denotes Layer Normalization~\cite{lei2016layer}.
From the $\hat{F}$, the LA aggregates neighborhood information of each patch token for locality awareness.
We utilize multiple convolutional layers $\{\mathrm{Conv}^{k_n}\}^{N_c}_{n=1}$ with different kernel size $k_n \times k_n$, where $k_n \in \mathbb{K}=\{k_1, k_2, ..., k_{N_c}\}$ to aggregate the neighborhood information.
The locality-aware feature $P$ is extracted as:
\begin{align} 
\label{eq:LA_conv}
 L^{k_n}  &= \mathrm{Conv}^{k_n}(\hat{F}), \\ 
 P &= \mathrm{FFN}(L^{k_1} + ... + L^{k_{N_{c}}}).
\end{align}
Then, $P$ is projected back to $D_{\mathrm{clip}}$ and fused with the original $F$ as follows:
\begin{align}
    F' &=  F + \gamma_{\mathrm{LA}} \cdot \mathrm{FFN}(P),
\end{align}
where $\gamma_{\mathrm{LA}}\in\mathbb{R}^{D_{\mathrm{clip}}}$ is a learnable parameter to balance between $P$ and $F$.

\subsection{Interaction Adapter}
Although the locality-aware feature helps the model capture fine-grained details of individual objects, it is insufficient to determine interactions, 
as interactions depend on the specific patterns between human and object contexts --specifically, how human cues are associated with object cues. 
For example, the `riding a bike' interaction is identified by recognizing the association between the human cues and the object cues such as the hands in contact with the handle.
This association distinguishes the interaction from other possible interactions, such as `repairing a bike.'
To enhance such interaction awareness, our Interaction Adapter (IA) updates each HO token based on its interaction pattern. 
We first extract region features for the human and object using ROIAlign~\cite{he2017mask}. 
These features are then refined by the Interaction Pattern Reasoning Module (IPRM) by capturing the interaction pattern.
The refined region features are subsequently used to inject interaction awareness into the corresponding HO token.

Specifically, the region features for the human and object in the $i$-th human-object pair are extracted as:
\begin{align} 
 R^{\tau}_{i}  &= \mathrm{FFN}(\mathrm{ROIAlign}(F',b^{\tau}_{i})),
\end{align}
where $\tau \!\in \!\{h, o\}$ is an indicator for human/object, and $b^{\tau}_i$ represents the corresponding bounding box of the $i$-th human-object pair.
Then, the IPRM captures human and object region contexts using learnable queries $Q\in \mathbb{R}^{N_p \times D_a}$ through a cross-attention mechanism: 
\begin{align} 
 \tilde{R}^{\tau}_{i} &= \mathrm{CrossAttn}(Q,R^{\tau}_{i},R^{\tau}_{i}).
\end{align}
We utilize the $N_p$ queries to capture the interaction-relevant contexts while filtering out irrelevant details.
Next, IPRM reasons about interaction patterns by computing how each region context is associated with its counterpart through cross-attention:
\begin{align} 
\label{eq:IA_human}
\hat{R}^{h}_{i} &= \mathrm{CrossAttn}(\tilde{R}^{h}_{i} ,\tilde{R}^{o}_{i},\tilde{R}^{o}_{i}), \\
\hat{R}^{o}_{i} &= \mathrm{CrossAttn}(\tilde{R}^{o}_{i} ,\tilde{R}^{h}_{i},\tilde{R}^{h}_{i}).
\label{eq:IA_object}
\end{align}
Subsequently, the HO token $T_i$ is projected into $D_a$ dimension through an FFN: $\tilde{T}_i = \mathrm{FFN}(T_i)$. $\tilde{T}_i$ is used as a query to extract interaction-aware features $\bar{R}^\tau_i$, which are utilized to update the HO token:
\begin{align} 
 \label{eq:IA_feat}
 \bar{R}^{\tau}_i &= \mathrm{CrossAttn}(\tilde{T}_i,\hat{R}^{\tau}_i,\hat{R}^{\tau}_i), \\
 T'_i &= T_i + \gamma_{\mathrm{IA}} \cdot \mathrm{FFN}([\bar{R}^{h}_i;\bar{R}^{o}_i]),
\end{align}
where $\gamma_{\mathrm{IA}}$ is a learnable parameter.
The updated HO tokens then are passed through the $l$-th layer of the CLIP visual encoder, replacing Eq.~\ref{eq:vit}:
\begin{equation}
    [T_{(l)}, \mathrm{cls}_{(l)}, F_{(l)}] = \mathcal{V}_{(l)}([T'_{(l-1)};\mathrm{cls}_{(l-1)};F'_{(l-1)}]).
\end{equation}

\subsection{Training and Inference}
Similar to the previous work~\cite{mao2024clip4hoi,wang2022learning}, we convert each HOI category into a text description using the template: ``A photo of a person [verb-ing] a [object]." 
We then insert several learnable tokens in front of the text description.
The text descriptions are fed into the CLIP text encoder to obtain the text embeddings $E\in\mathbb{R}^{N_{|\mathbb{C}|} \times D_{\mathrm{clip}}}$ for all HOI categories, where $N_{|\mathbb{C}|}$ denotes the number of HOI categories. After obtaining the text embeddings, the HOI scores $S\in\mathbb{R}^{N_{\mathrm{pair}} \times N_{|\mathbb{C}|}}$ can be calculated as:
\begin{align}
 S = \mathrm{Sigmoid}(T_{(L)}E^\top / \tau),
\end{align}
where $\tau$ is the learnable parameter for rescaling the logits.
Since a human can engage in multiple interactions with an object, we utilize the sigmoid function instead of softmax to compute the HOI scores.

\smallskip
\noindent \textbf{Training.} 
To train our proposed method, we assign positive labels to samples whose human and object bounding boxes both have an Intersection-over-Union (IoU) exceeding a threshold with the ground truth.
Following the previous work~\cite{mao2024clip4hoi,lei2023ada}, we adopt the binary focal loss~\cite{lin2017focal}:
\begin{align}
 \mathcal{L} = \mathrm{FocalBCE}(S,Y),
\end{align}
where $Y\in\{0,1\}^{N_\mathrm{pair}\times N_{|\mathcal{C}|}}$ represents the binary target labels.

\smallskip
\noindent \textbf{Inference.} 
During inference, we incorporate the confidence scores of the human and object boxes from DETR into the HOI scores as:
\begin{align}
 S_\mathrm{infer} = S \cdot S^\lambda_{H} \cdot S^\lambda_{O},
\end{align}
where $\lambda$ is a hyper-parameter for suppressing overconfident detections~\cite{zhang2022upt,zhang2021spatially}.
$S_H, S_O\in[0,1]^{N_\mathrm{pair}\times 1}$ denotes the confidence scores of human and object for corresponding human-object pairs, respectively.

%% file: sections/4_experiments.tex
\section{Experiments}
\subsection{Experiment Settings}

To show the effectiveness of the proposed method, we evaluate our model on the two public benchmark datasets: HICO-DET~\cite{hico} and V-COCO~\cite{vcoco}.

\begin{table*}[t!]
\resizebox{0.98\linewidth}{!}{
\begin{tabular}{c ccc ccc ccc ccc ccc}
\toprule
\multirow{2}{*}{Method} & \multicolumn{3}{c}{RF-UC} & \multicolumn{3}{c}{\cellcolor{gray!20}NF-UC} & \multicolumn{3}{c}{UO} & \multicolumn{3}{c}{\cellcolor{gray!20}UV} & \multicolumn{3}{c}{UC} \\
\cmidrule{2-4} \cmidrule{5-7} \cmidrule{8-10} \cmidrule{11-13} \cmidrule{14-16}

& Unseen & Seen & Full & \cellcolor{gray!20}Unseen & \cellcolor{gray!20}Seen & \cellcolor{gray!20}Full & Unseen & Seen & Full & \cellcolor{gray!20}Unseen & \cellcolor{gray!20}Seen & \cellcolor{gray!20}Full & Unseen & Seen & Full \\
\midrule
FCL~\cite{hou2021detecting} & 13.16 & 24.23 & 22.01 & \cellcolor{gray!20}18.66 & \cellcolor{gray!20}19.55 & \cellcolor{gray!20}19.37 & 15.54 & 20.74 & 19.87 & \cellcolor{gray!20}- & \cellcolor{gray!20}- & \cellcolor{gray!20}- & - & - & - \\
ATL~\cite{hou2021affordance} & 9.18 & 24.67 & 21.57 & \cellcolor{gray!20}18.25 & \cellcolor{gray!20}18.78 & \cellcolor{gray!20}18.67 & 15.11 & 21.54 & 20.47 & \cellcolor{gray!20}- & \cellcolor{gray!20}- & \cellcolor{gray!20}- & - & - & - \\
RLIP~\cite{yuan2022rlip} & 19.19 & 33.35 & 30.52 & \cellcolor{gray!20}20.27 & \cellcolor{gray!20}27.67 & \cellcolor{gray!20}26.19 & - & - & - & \cellcolor{gray!20}- & \cellcolor{gray!20}- & \cellcolor{gray!20}- & - & - & - \\
GEN-VLKT~\cite{liao2022gen} & 21.36 & 32.91 & 30.56 & \cellcolor{gray!20}25.05 & \cellcolor{gray!20}23.38 & \cellcolor{gray!20}23.71 & 10.51 & 28.92 & 25.63 & \cellcolor{gray!20}20.96 & \cellcolor{gray!20}30.23 & \cellcolor{gray!20}28.74 & - & - & - \\
LOGICHOI~\cite{li2024logic} & 25.97 & 34.93 & 33.17 & \cellcolor{gray!20}26.84 & \cellcolor{gray!20}27.86 & \cellcolor{gray!20}27.95 & 15.67 & 30.42 & 28.23 & \cellcolor{gray!20}- & \cellcolor{gray!20}- & \cellcolor{gray!20}- & - & - & - \\
ADA-CM~\cite{lei2023ada} & 27.63 & 34.35 & 33.01 & \cellcolor{gray!20}32.41 & \cellcolor{gray!20}31.13 & \cellcolor{gray!20}31.39 & - & - & - & \cellcolor{gray!20}- & \cellcolor{gray!20}- & \cellcolor{gray!20}- & - & - & - \\
EoID~\cite{wu2023eoid} & 22.04 & 31.39 & 29.52 & \cellcolor{gray!20}26.77 & \cellcolor{gray!20}26.66 & \cellcolor{gray!20}26.69 & - & - & - & \cellcolor{gray!20}22.71 & \cellcolor{gray!20}30.73 & \cellcolor{gray!20}29.61 & 23.01 & 30.39 & 28.91 \\
HOICLIP~\cite{ning2023hoiclip} & 25.53 & 34.85 & 32.99 & \cellcolor{gray!20}26.39 & \cellcolor{gray!20}28.10 & \cellcolor{gray!20}27.75 & 16.20 & 30.99 & 28.53 & \cellcolor{gray!20}24.30 & \cellcolor{gray!20}32.19 & \cellcolor{gray!20}31.09 & 23.15 & 31.65 & 29.93 \\
CLIP~\cite{CLIP} & 28.79 & 22.00 & 23.36 & \cellcolor{gray!20}28.52 & \cellcolor{gray!20}22.06 & \cellcolor{gray!20}23.36 & 28.66 & 22.29 & 23.36 & \cellcolor{gray!20}26.16 & \cellcolor{gray!20}22.90 & \cellcolor{gray!20}23.36 & 24.28 & 23.12 & 23.36 \\
CLIP4HOI~\cite{mao2024clip4hoi} & 28.47 & 35.48 & 34.08 & \cellcolor{gray!20}31.44 & \cellcolor{gray!20}28.26 & \cellcolor{gray!20}28.90 & 31.79 & 32.73 & 32.58 & \cellcolor{gray!20}26.02 & \cellcolor{gray!20}31.14 & \cellcolor{gray!20}30.42 & 27.71 & 33.25 & 32.11 \\
BCOM$^{\dagger}$~\cite{wang2024bilateral} & 28.52 & 35.04 & 33.74 & \cellcolor{gray!20}33.12 &  \cellcolor{gray!20}31.76 &  \cellcolor{gray!20}32.03 &  - & - & - & \cellcolor{gray!20}- & \cellcolor{gray!20}- &\cellcolor{gray!20}- & - & - & - \\
CMMP~\cite{lei2025exploring} & 29.45 & 32.87 & 32.18 & \cellcolor{gray!20}32.09 & \cellcolor{gray!20}29.71 & \cellcolor{gray!20}30.18 & 33.76 & 31.15 & 31.59 & \cellcolor{gray!20}26.23 & \cellcolor{gray!20}32.75 & \cellcolor{gray!20}31.84 & 29.60 & 32.39 & 31.84 \\
\midrule
\textbf{\ours}$^{\phantom{\dagger}}$ & \textbf{31.83} & \textbf{35.06} & \textbf{34.41} & \cellcolor{gray!20}\textbf{36.41} & \cellcolor{gray!20}\textbf{32.44} & \cellcolor{gray!20}\textbf{33.23} & \textbf{37.88} & \textbf{33.55} & \textbf{34.27} & \cellcolor{gray!20}\textbf{28.96} & \cellcolor{gray!20}\textbf{33.80} & \cellcolor{gray!20}\textbf{33.12} & \textbf{31.64} & \textbf{35.04} & \textbf{34.36} \\

\textbf{\ours}$^{\dagger}$ & \textbf{36.57} & \textbf{38.54} & \textbf{38.13} & \cellcolor{gray!20}\textbf{37.52} & \cellcolor{gray!20}\textbf{35.90} & \cellcolor{gray!20}\textbf{36.22} &  \textbf{40.78} & \textbf{36.96} & \textbf{37.60} & \cellcolor{gray!20}\textbf{32.05} & \cellcolor{gray!20}\textbf{38.04} & \cellcolor{gray!20}\textbf{37.20} &

\textbf{32.25} & \textbf{37.95} & \textbf{36.81}

\\
\bottomrule
\end{tabular}
}
\vspace{-2mm}

\caption{Performance comparison on the HICO-DET dataset under various zero-shot settings. RF-UC, NF-UC, UO, UV, and UC denote rare first unseen composition, non-rare first unseen composition, unseen object, unseen verb, and unseen composition settings, respectively. Our method outperforms existing methods, demonstrating the effectiveness of the proposed methods. The highest result in each section is highlighted in bold. $\dagger$ indicates CLIP with ViT-L backbone.}
\label{tab:zero_shot}
\vspace{-4mm}
\end{table*}

\textbf{HICO-DET} has 38,118 images for training and 9,658 images for testing. 
It contains 80 object classes, 117 interaction classes, and 600 HOI categories.
Following conventional evaluation protocol~\cite{mao2024clip4hoi,lei2023ada,ning2023hoiclip}, we report the mean average precision (mAP) to examine the model performance on five zero-shot settings: Unseen Combination(UC), Rare First Unseen Combination (RF-UC), Non-rare First Unseen Combination (NF-UC), Unseen Verb (UV), Unseen Object (UO). 
In the UC setting, all object and verb categories appear during the training; however, some HOI categories do not appear during the training, and they are used as $\mathbb{C}^{\mathrm{UC}}_{\mathrm{unseen}}$.
Especially the least frequent 120 HOI categories are used as $\mathbb{C}^{\mathrm{RF-UC}}_{\mathrm{unseen}}$ while  the most frequent 120 HOI categories are used as $\mathbb{C}^{\mathrm{NF-UC}}_{\mathrm{unseen}}$. 
In the UV setting, 20 verb categories ($\mathbb{V}_{\mathrm{unseen}}$) are not used during the training, and corresponding HOI categories are used as unseen HOI categories, \textit{i.e.}, $\mathbb{C}^{\mathrm{UV}}_{\mathrm{unseen}}=\{(o_i, v_j)|o_i \in \mathbb{O}, v_j \in \mathbb{V}_{\mathrm{unseen}}\}$. 
Similarly, in the UO setting, 12 object categories ($\mathbb{O}_{\mathrm{unseen}}$) are not used during the training, and corresponding HOI categories are used as unseen HOI categories, \textit{i.e.}, $\mathbb{C}^{\mathrm{UO}}_{\mathrm{unseen}} = \{(o_i,v_j)|o_i \in \mathbb{O}_{\mathrm{unseen}},v_j \in \mathbb{V}\}$. 

\textbf{V-COCO} is a subset of the MS-COCO~\cite{coco} dataset. It consists of 5,400 and 4,946 images for training and testing.
V-COCO consists of 80 object classes and 29 action classes. 
Following evaluation settings in ~\cite{kim2021hotr}, we evaluate \ours~on scenario 2, and report role average precision $\mathrm{AP}^{\#2}_{\mathrm{role}}$.

\subsection{Comparison with State-of-the-Art Methods}
\noindent\textbf{Zero-shot settings.} 
We evaluate the performance of \ours~and compare it with existing HOI detection methods under various zero-shot settings.
As shown in Table~\ref{tab:zero_shot}, \ours~demonstrates effectiveness by outperforming all previous methods by a significant margin under all zero-shot settings.
In particular, existing methods~\cite{lei2023ada,mao2024clip4hoi,ning2023hoiclip,wu2023eoid,wang2024bilateral,lei2025exploring} that leverage CLIP representation show lower or comparable performance on unseen classes than CLIP itself under the RF-UC and UV settings.
These results indicate that adapting CLIP representation for zero-shot HOI detection weakens its generalization ability due to the domain gap between the image-level pre-training task and HOI detection.
In contrast, our model consistently outperforms CLIP and other existing methods on unseen classes, demonstrating its effectiveness.
Furthermore, when we increase the model size to ViT-L, \ie, LAIN$^\dagger$, the performance further improves, demonstrating the scalability of the proposed method with a larger backbone.
Notably, despite BCOM$^{\dagger}$~\cite{wang2024bilateral} using the larger ViT-L backbone, \ours still surpasses it by a significant margin using only ViT-B.
These results indicate that it is crucial for CLIP representation to consider fine-grained details of individual objects and interaction patterns between humans and objects.

\noindent\textbf{Fully-supervised settings.}
To further validate the effectiveness of our proposed method, we conducted experiments under conventional fully-supervised settings on the HICO-DET and V-COCO datasets. 
As shown in Table~\ref{tab:full_sup}, on the HICO-DET dataset, \ours not only surpasses both fully-supervised models and zero-shot methods but also shows a marked improvement on rare HOI classes, which present significant challenges due to their scarcity and difficulty in generalizing, similar to unseen classes.
Despite having fewer parameters and FLOPS, as shown in Table~\ref{tab:flops},~\ours~demonstrates competitive performance on the V-COCO dataset, achieving the second-best results among zero-shot methods.

\begin{table}[t!]

\resizebox{\linewidth}{!}{%
\begin{tabular}{c ccc c}
\toprule
\multirow{2}{*}{Method} & \multicolumn{3}{c}{HICO-DET} & V-COCO              \\
              &           Full & Rare   & Non-rare   & $\mathrm{AP}^{\#2}_{\mathrm{role}}$ \\
\midrule
\textit{\textbf{Fully-supervised methods}} \\
HOTR~\cite{kim2021hotr}  & 25.10 & 17.34 & 27.42 & 64.4 \\
ATL~\cite{hou2021affordance}  & 28.53 & 21.63 & 30.59  & -  \\
As-Net~\cite{chen2021asnet}  & 28.87 & 24.25 & 30.25  & -  \\
QPIC~\cite{tamura2021qpic}  & 29.07 & 21.85 & 31.23 & 61.0 \\
UPT~\cite{zhang2022upt}  & 31.66 & 25.94 & 33.36 & 64.5 \\
CDN~\cite{zhang2021cdn} & 31.78 & 27.55 & 33.05 & 64.4 \\
Iwin~\cite{tu2022iwin} & 32.03 & 27.62 & 34.14 & 60.5 \\
GEN-VLKT~\cite{liao2022gen} & 33.75 & 29.25 & 35.10 & 64.5 \\
ADA-CM~\cite{lei2023ada} & 33.80 & 31.72 & 34.42 & 61.2\\
LogicHOI~\cite{li2024logic}  & 35.47 & 32.03 & \textbf{36.22} & \underline{65.6} \\

\textit{\textbf{Zero-shot methods}} \\
HOICLIP~\cite{ning2023hoiclip} & 34.69 & 31.12 & 35.74 & 64.8 \\
CLIP4HOI~\cite{mao2024clip4hoi}& 35.33 & 33.95 & 35.74 & \textbf{66.3} \\
CMMP~\cite{lei2025exploring} & 32.26 & 33.53 & 33.24 & 61.2\\
\midrule

\ours & \textbf{36.02} & \textbf{35.70} & \underline{36.11} & 65.1 \\

\bottomrule
\end{tabular}
}
\vspace{-2mm}
\caption{Performance comparison on the HICO-DET~\cite{hico} and V-COCO~\cite{vcoco} datasets under fully-supervised setting. The highest result in
each section is highlighted in bold.}
\label{tab:full_sup}
\vspace{-4mm}
\end{table}

\subsection{Ablation Study}
We conduct various ablation studies on the UV setting to validate the effectiveness of \ours.

\noindent\textbf{The impact of each adapter.} 
In Table~\ref{tab:ablation_compo}, we gradually add each adapter to the baseline, which utilizes the original CLIP representation without LA and IA, to investigate the impact of each adapter. 
We observed performance improvements in both seen and unseen classes by injecting locality awareness into CLIP representations through the LA. 
This result indicates that capturing the fine-grained information of individual objects is crucial for zero-shot HOI detection.
Furthermore, applying IA to enhance interaction awareness in CLIP representations leads to performance improvements, emphasizing the importance of understanding interaction patterns between humans and objects.
Moreover, since fine-grained information about individual objects aids in reasoning about interaction patterns, we observed that using both adapters together yielded the best performance. 
This demonstrates the effectiveness of both adapters, \ie, LA and IA, and highlights the importance of incorporating locality and interaction awareness for adapting CLIP representations to zero-shot HOI detection.

\begin{table}[t!]
    \centering
    \resizebox{0.65\columnwidth}{!}{%
    \begin{tabular}[t]{ccccc}
    \toprule
     LA & IA & Unseen & Seen & Full \\
     \midrule
     - & - & 24.88 & 31.06 & 30.19 \\
     \checkmark & - & 27.71 & 32.55 & 31.95 \\
     - & \checkmark & 27.37 & 33.57 & 32.70  \\
     \checkmark  & \checkmark & \textbf{30.50} & \textbf{34.80} & \textbf{33.95} \\
     \bottomrule
    \end{tabular}
       }
    \vspace{-2mm}
    \caption{Ablations studies on each adapter under UV setting. LA and IA denote locality and interaction adapter, respectively.}
    \label{tab:ablation_compo}
    \vspace{-6mm}
    
\end{table}

\noindent\textbf{The impact of each component in LA.} To investigate the impact of each component in the LA, we conduct comparisons between the full LA model and various LA variants in Table~\ref{tab:ablation_design} (a) to (d).
Removing either the visual context or the spatial layout component, \ie, (a) or (b), results in a significant decline in performance. This demonstrates that integrating both surrounding visual context and spatial layout is essential for capturing fine-grained details of individual objects.
Furthermore, we replace our convolutional layers with existing attention mechanisms~\cite{ramachandran2019stand,liu2021swin}, which are designed to capture local information, \ie, fine-grained details, instead of global information, as shown in (c) and (d).
We observe that the existing mechanisms improve performance on both unseen and seen classes compared to baseline, \ie, without LA and IA, by capturing local information.
These results demonstrate the importance of fine-grained details about individual objects in adapting CLIP’s representations for zero-shot HOI detection.
However, their performances are degraded compared to our LA.
This highlights the effectiveness of LA in capturing fine-grained details.

\noindent\textbf{The impact of each component in IA.}
We also investigate the impact of each component in IA on achieving interaction awareness. In Table~\ref{tab:ablation_design}, we compare the full IA model with various IA variants.
When we remove the IPRM, the model's performance on unseen classes significantly degrades, indicating that incorporating the interaction pattern is crucial for interaction awareness as shown in (e). 
Similarly, removing human/object context extraction (\ie, extracting the interaction pattern using only ROI-aligned features) results in decreased model performance, suggesting that capturing interaction-relevant contexts while filtering out irrelevant details through context extraction is effective for reasoning about the interaction pattern as shown in (f).

\begin{table}[t!]
    \centering
    \resizebox{0.83\columnwidth}{!}{%
    \begin{tabular}[t]{cccccc}
    \toprule
      &Method & Unseen & Seen & Full \\
     \midrule
     & Baseline &24.88 & 31.06 & 30.19 \\
     \midrule
     &\textit{\textbf{Locality Adapter}} & \textbf{27.71} & \textbf{32.55} & \textbf{31.95} \\
    (a) &w.o visual information & 26.77 & 32.18 & 31.40\\
    (b) &w.o spatial layout & 26.52 & 32.07 & 31.31 \\
    (c) &Local Attention~\cite{ramachandran2019stand} & 26.46 & 32.39  & 31.56\\
    (d) &Window Attention~\cite{liu2021swin} & 26.35 & 32.31 & 31.47 \\
     \midrule
     &\textit{\textbf{Interaction Adapter}} & \textbf{27.37} & \textbf{33.57} & \textbf{32.70} \\
       (e)  & w.o IPRM & 24.32 & 32.76 & 31.57\\
   (f) &w.o human/object context &  25.64 & 32.41 & 31.40 \\

     \bottomrule
    \end{tabular}
       }
       \vspace{-2mm}
            \caption{Ablation study on the design choices of each adapter under UV setting. }
    \vspace{-2mm}
     \label{tab:ablation_design}
\end{table}

\begin{table}[t!]
    \centering
    \resizebox{0.94\linewidth}{!}{
        \begin{tabular}{ccccccccc}
        \toprule
        & \multirow{2}{*}{\textbf{Method}} & \multicolumn{3}{c}{\textbf{Unseen}} & \multicolumn{3}{c}{\textbf{Seen}} \\ \cmidrule(lr){3-5} \cmidrule(lr){6-8}
        & & \textbf{AP$_{L}$} & \textbf{AP$_{M}$} & \textbf{AP$_{S}$} & \textbf{AP$_{L}$} & \textbf{AP$_{M}$} & \textbf{AP$_{S}$} \\ 
        \midrule
        \multirow{3}{*}{\rotatebox{90}{Human}} 
        & CMMP~\cite{lei2025exploring} & 45.77 & 34.54 & 52.89 & 64.80 & 54.60 & 25.05 \\
        & ADA-CM~\cite{lei2023ada} & \textbf{50.63} & 36.85 & 26.88 & 60.80 & 39.13 & 17.57 \\
        & \ours & \underline{48.11} & \textbf{38.66} & \textbf{62.32} & \textbf{66.51} & \textbf{59.92} & \textbf{30.44} \\ 
        \midrule\multirow{3}{*}{\rotatebox{90}{Object}} 
        & CMMP~\cite{lei2025exploring} & 59.64& 46.19 & 18.64 & 62.73 & 51.38 & 29.82 \\
        & ADA-CM~\cite{lei2023ada} & \textbf{62.16} & 46.60 & 14.02 & 57.20 & 37.99 & 18.66 \\
        & \ours & \underline{61.41} & \textbf{49.45} & \textbf{20.82} &  \textbf{63.82} & \textbf{56.93} & \textbf{35.91} \\ 
        \bottomrule
        \end{tabular}
    }
    \vspace{-2mm}
    \caption{Comparison of AP across different box sizes under the RF-UC setting.\protect\footnotemark}
    \vspace{-2mm}
    \label{tab:ap_comparison}
\end{table}

\begin{table}
    \centering
    \resizebox{0.67\columnwidth}{!}{%
        \begin{tabular}{cccc}
            \toprule
            position & Unseen & Seen & Full \\
            \midrule
            $\emptyset$ & 24.88 & 31.06 & 30.19\\
            1-6 & 27.29 & 32.11 & 31.46 \\
            7-12 & 27.83 & 32.72 & 32.04 \\ 
            1-12 & \textbf{28.96} & \textbf{33.80} & \textbf{33.12}\\
            \bottomrule
        \end{tabular}
    }
    \vspace{-2mm}

    \caption{Ablation study evaluating the impact of different adapter positions in the UV setting.}
    \vspace{-5mm}
    \label{tab:position}
\end{table}

\footnotetext{Since the authors of ADA-CM~\cite{lei2023ada} provide pretrained weight under UC-RF instead of UV, we conduct our experiments under the UC-RF.}

\noindent\textbf{The impact of locality-aware adaptation.}
To validate that our IA and LA modules enhance CLIP's ability to focus on local details, we conduct experiments varying human and object box sizes, as local details become more critical with smaller boxes.
As shown in Table~\ref{tab:ap_comparison}, as the box size decreases from large (AP$_L$) to small (AP$_S$), LAIN demonstrates a larger performance gap for medium and small boxes compared to large ones.
Specifically, \ours outperforms ADA-CM~\cite{lei2023ada} by 4.91\% (human) and 6.12\% (object) in AP$_M$, and notably by 131.85\% (human) and 48.50\% (object) in AP$_S$ under the unseen setting. The performance gain becomes increasingly prominent as box sizes decrease (M→S), and similar results are also observed in the seen classes.
These results validate the effectiveness of our IA and LA in capturing fine-grained local details for zero-shot HOI detection.

\noindent\textbf{The impact of adapter positioning.}
In Table~\ref{tab:position}, we conduct experiments that explore the impact of adapter positioning.
We divide the model layers into lower and upper halves, inserting adapters into each section separately.
We observe that adding adapters to either the lower layers (\ie, 1-6) or the upper layers (\ie, 7-12) consistently improves the model's performance.
In particular, we observe that incorporating adapters into the upper layers results in a more substantial performance gain since the upper layers in the ViT structure encode global information, unlike the lower layer which encodes local information~\cite{dosovitskiy2020image}.
These results suggest that injecting interaction and locality awareness into features lacking local information is essential and that our proposed method effectively incorporates interaction and locality awareness into CLIP representations.

\noindent\textbf{Parameter analysis.} 
In Table~\ref{tab:flops}, we provide a comparative analysis of model parameters and computational cost between our proposed method and state-of-the-art methods for zero-shot HOI detection to show the efficiency of ~\ours.
Compared to HOICLIP~\cite{ning2023hoiclip} and CLIP4HOI~\cite{mao2024clip4hoi}, which introduce a large number of trainable parameters and FLOPs as they utilize a heavy decoder for interaction classification, \ours introduces only 3.0M trainable parameters, alongside reduced FLOPs of 110G.
Although \ours has more trainable parameters than CMMP~\cite{lei2025exploring}, it has fewer total parameters and FLOPs.
This highlights the efficiency of our approach in incorporating locality and interaction awareness with minimal added overhead.

\begin{figure}[t!]
    \centering
    \resizebox{0.98\linewidth}{!}{\includegraphics[width=\linewidth]{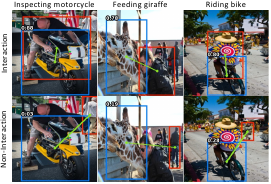}}
    \vspace{-3mm}
    \caption{Qualitative results on HICO-DET under UV settings.
    We represent a human with a red box and an object with a blue box, along with HOI score.}
    \label{fig:qual}
    \vspace{-6mm}
\end{figure}

\subsection{Qualitative Results}

\begin{figure}[t!]
    \centering
     \resizebox{0.98\linewidth}{!}{\includegraphics[width=\linewidth]{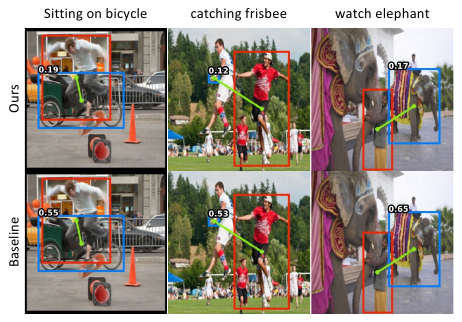}}
    \vspace{-3mm}
    \caption{Qualitative comparison of non-interactive pairs between \ours and the baseline, \ie, without LA and IA, on the HICO-DET under the UV setting. We represent a human with a red box and an object with a blue box, along with HOI scores.}
    \label{fig:qual2}
    \vspace{-2mm}
\end{figure}

We present qualitative results on the HICO-DET under the UV setting in Figure~\ref{fig:qual} and \ref{fig:qual2}. 
In Figure~\ref{fig:qual}, we observe that \ours successfully distinguishes interactive pairs, assigning high similarity scores, and non-interactive pairs, assigning low similarity scores, even for unseen verbs, \ie, `inspecting', `feeding', and `riding'.
Moreover, as shown in Figure~\ref{fig:qual2}, \ours assigns significantly lower scores to non-interactive pairs compared to the baseline, which does not incorporate LA and IA.
The results demonstrate that incorporating LA and IA leads to more discriminative HOI representations, enabling the model to better distinguish interactive and non-interactive pairs for zero-shot HOI detection.

\begin{table}
    \centering
        \resizebox{0.8\columnwidth}{!}{%
        \begin{tabular}[t]{cccc}
    \toprule
     Method &  Tr. Params  & Tot. Params  & FLOPs \\
     \midrule 
     HOICLIP~\cite{ning2023hoiclip} & 193.3M & 193.3M & 179G  \\
     CLIP4HOI~\cite{mao2024clip4hoi} & 71.2M & 262.4M & 186G \\
     CMMP~\cite{lei2025exploring} & \textbf{2.3M} & 193.4M & 114G \\
     \ours & 3.0M & \textbf{145.4M} & \textbf{110G}\\
     \bottomrule
    \end{tabular}
    }
    \vspace{-2mm}
\caption{Comparison of parameters across state-of-the-art models under UV setting. Tr. and Tot. Params represent the number of trainable and total parameters of the model, respectively.}
    \label{tab:flops}
    \vspace{-6mm}
\end{table}

%% file: sections/5_conclusion.tex
\section{Conclusion}

In this paper, we have proposed \ours, designed to address the lack of local details in CLIP's representation, which hinders CLIP's generalization ability when adapting to zero-shot HOI detection.
Our locality adapter introduces locality awareness into CLIP by considering surrounding visual information and spatial layout.
For interaction awareness, which is difficult to determine solely through locality awareness, our interaction adapter infers the interaction pattern by leveraging contextual reasoning between human and object contexts.
By enhancing the locality and interaction awareness, \ours~effectively captures fine-grained information about HO pairs, facilitating adaptation of CLIP's representation to zero-shot HOI detection.
Extensive experiments on two public benchmarks, HICO-DET and V-COCO, demonstrate the importance of capturing local details and the effectiveness of \ours.
Notably, \ours is particularly effective in scenarios where local details are crucial, \ie, small instances, while introducing minimal computational cost.

\noindent \textbf{Acknowledgements.}
This work was supported by the NRF grant (RS-2021-NR059830 (50\%)) and the IITP grants (RS-2022-II220959: Few-Shot Learning of Causal Inference in Vision and Language for Decision Making (45\%), RS-2019-II191906: AI Graduate School Program at POSTECH (5\%)) funded by the Korea government (MSIT).